\documentclass[letterpaper]{article} 
\usepackage{aaai2026}  
\usepackage{times}  
\usepackage{tikz}
\usepackage{tabularx}
\usepackage{booktabs}
\usepackage{helvet}  
\usepackage{courier}  
\usepackage[hyphens]{url}  
\usepackage{graphicx} 
\usepackage{natbib}  
\usepackage{caption} 
\usepackage{algorithm}
\usepackage{algorithmic}
\usepackage{overpic}
\usepackage{multirow}
\usepackage{amsmath}
\usepackage{amsfonts} 
\usepackage{mathrsfs}
\usepackage{newfloat}
\usepackage{listings}
\DeclareCaptionStyle{ruled}{labelfont=normalfont,labelsep=colon,strut=off} 
\floatstyle{ruled}
\newfloat{listing}{tb}{lst}{}
\floatname{listing}{Listing}
\title{TOSC: Task-Oriented Shape Completion for Open-World Dexterous Grasp Generation from Partial Point Clouds}
\author{
    Weishang Wu\textsuperscript{\rm 1}, Yifei Shi\textsuperscript{\rm 1}\thanks{Corresponding author: Yifei Shi, Zhiping Cai}, Zhiping Cai\textsuperscript{\rm 1}\footnotemark[1]
}
\affiliations{
    \textsuperscript{\rm 1}National University of Defense Technology\\
    wuweishang24@nudt.edu.cn, yifei.j.shi@gmail.com, zpcai@nudt.edu.cn
}

\usepackage{bibentry}

\begin{document}

\maketitle

\begin{abstract}
Task-oriented dexterous grasping remains challenging in robotic manipulations of open-world objects under severe partial observation, where significant missing data invalidates generic shape completion. In this paper, to overcome this limitation, we study \emph{Task-Oriented Shape Completion}, a new task that focuses on completing the potential contact regions rather than the entire shape. We argue that shape completion for grasping should be explicitly guided by the downstream manipulation task. To achieve this, we first generate multiple task-oriented shape completion candidates by leveraging the zero-shot capabilities of object functional understanding from several pre-trained foundation models. A 3D discriminative autoencoder is then proposed to evaluate the plausibility of each generated candidate and optimize the most plausible one from a global perspective. A conditional flow-matching model named FlowGrasp is developed to generate task-oriented dexterous grasps from the optimized shape. Our method achieves state-of-the-art performance in task-oriented dexterous grasping and task-oriented shape completion, improving the Grasp Displacement and the Chamfer Distance over the state-of-the-art by $16.17\%$ and $55.26\%$, respectively. In particular, it shows good capabilities in grasping objects with severe missing data. It also demonstrates good generality in handling open-set categories and tasks. 

\end{abstract}
\begin{links}
    \link{Code}{https://github.com/SyKszzzzz/TOSC}
\end{links}

\section{Introduction}
\label{sec:intro}
Generating dexterous grasps that are both stable and effective for specific downstream tasks remains a core challenge in robot manipulation, attracting substantial research attention. Task-oriented dexterous grasping addresses this challenge by generating grasp poses explicitly designed to facilitate the intended manipulation~\cite{zhong2025dexgraspvla, wei2024grasp}. This capability is essential for advancing versatile robotic applications in fields such as household service and industrial automation~\cite{she2024learning, liu2025prevalence}.

Recent advances in large foundation models have enabled zero-shot task-oriented dexterous grasping for open-world objects. 
While existing methods show promise on synthetic objects or real objects with complete geometry~\cite{li2024semgrasp, zhong2025dexgraspvla, li2024multi, mirjalili2024lan, jian2025g}, their performance significantly deteriorates in cluttered real-world environments, due to the influences of severe occlusion, background clutter, and sensor noise.
This limitation essentially stems from the requirements of detailed functional understanding, which are unattainable without knowing the completed geometry.

A straightforward solution to address this problem is first to perform a shape completion on the input data, then generate grasps based on the completed shape~\cite{iwase2025zerograsp, kim2025dreamgrasp}. However, this decoupled approach would yield incorrect estimation of shape geometry for both the entire object and its contact regions, primarily due to the
ambiguity caused by data missing. Consequently, grasps generated on these erroneous completions might fail to satisfy the requirements of the specific grasp tasks.

\begin{figure}[t]\centering
    \begin{overpic}[width=0.95\linewidth,tics=10]{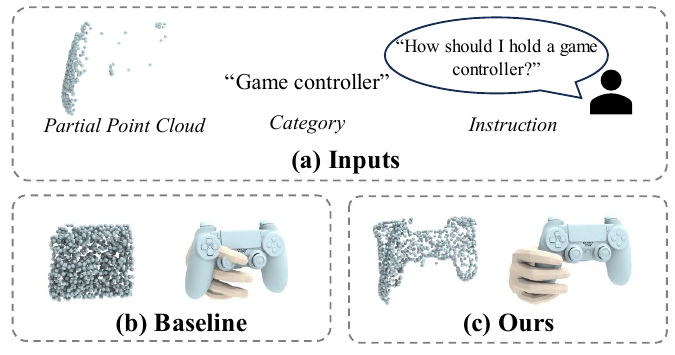}
    \end{overpic}
    \caption{By targeting task-oriented shape completion instead of generic shape completion, our method achieves higher completion accuracy, enabling more plausible task-oriented grasps, compared to the baseline~\cite{wei2024grasp}.}
    \label{fig:teaser}
\end{figure}

To solve this problem, this paper studies \emph{Task-Oriented Shape Completion}, a new task focusing exclusively on reconstructing contact regions rather than full object geometry. Our core insight is that shape completion for grasping should be explicitly conditioned on the downstream manipulation task (see Figure~\ref{fig:teaser}). As a consequence, the completion process is task-aware and capable of generating completed shapes that facilitate the execution of the specific manipulation task, tolerating the imperfections in irrelevant regions.

To this end, we propose a method that consists of several key components. First, it generates multiple candidates of task-oriented completed shapes by leveraging the zero-shot capabilities of object functional understanding from pre-trained foundation models. Specifically, it first synthesizes multiple plausible RGB images containing the potential contact regions conditioned on the input point cloud through the ControlNet~\cite{zhang2023adding}. It then generates 3D shapes from the RGB images by adopting a 3D shape generation network~\cite{zhao2025hunyuan3d}.

Second, the generated shapes may be imperfect due to hallucinations during RGB image synthesis and incorrect estimations during shape generation. To alleviate this problem, we propose a 3D discriminative autoencoder (DAE) to select the optimal generated shape and further optimize its geometry from a global perspective. The 3D DAE is trained on large-scale datasets with a well-designed training data generation scheme, enabling it to accurately restore the shape of open-set categories.

Third, we proposed FlowGrasp, a conditional flow-matching model to generate task-oriented dexterous grasps from the optimal shape.
Specifically, in the standard flow-matching framework, we apply a single-step, input-side gradient correction to each predicted velocity to implicitly enforce geometric and semantic constraints. Then merge the corrected velocity with the original flow target to update the model. This requires no additional explicit losses or inference overhead, yet guides the model toward constraint-aware optimization.

We conduct extensive experiments to evaluate the effectiveness of the proposed method. Our method achieves state-of-the-art performance in task-oriented dexterous grasping and task-oriented shape completion, improving the Grasp Displacement and the Chamfer Distance over the state-of-the-art by $16.17\%$ and $55.26\%$, respectively. In particular, it shows good capabilities in grasping objects with severe missing data. It also demonstrates good generality in handling open-set categories and tasks.

In summary, the contributions of this paper are as follows:
\begin{itemize}
     \item We propose task-oriented shape completion, a new task that focuses exclusively on completing contact regions rather than the full object geometry.
     \item We develop a method that first generates multiple task-oriented shape completion candidates by several pre-trained foundation models and then selects the most plausible one as well as optimizes its geometry from a global perspective by developing a 3D DAE.
     \item We introduce a constraint-aware conditional flow-matching model that enforces geometric and semantic constraints via a single-step gradient correction, with no extra losses or inference overhead.
     \item Our method achieves state-of-the-art performance on task-oriented dexterous grasping and shape completion.
\end{itemize}
\label{sec:Related Work}
\section{Related Work}
\subsection{Point Cloud Completion}
Point cloud completion is a long-standing research topic. Recent advances in point cloud completion are dominated by learning-based methods. Most methods directly estimate the missing part in a single forward pass by training on a large number of labeled data~\cite{yuan2018pcn,yu2021pointr}. Despite the satisfactory performance, the performance of these methods relies heavily on the quality of training data~\cite{li2025genpc} and has limited capability in out-of-distribution data. Another line of work integrates external 2D/3D priors to improve the generality~\cite{kasten2023point,huangcompc,li2025genpc}. Our method is inspired by the existing methods. However, our method explicitly incorporates the manipulation task to guide the completion process, making it a more practicable solution for robot manipulation.

\subsection{Task-Oriented Grasping}
Task-oriented grasping is a crucial yet challenging task that requires generating grasps not only for stability but also to enable the execution of a subsequent task. Existing methods aggregate features from language and vision inputs, providing zero-shot capability to both novel instruction and category~\cite{wei2024grasp,li2024semgrasp}. For example, DexTOG~\cite{zhang2024dextog} learns intermediate features for grasp-instruction alignment. DexGraspVLA~\cite{zhong2025dexgraspvla} proposes a hierarchical vision-language-action framework to implement grasp generation. AffordDexGrasp~\cite{wei2025afforddexgrasp} proposes a generalizable affordance representation that aligns robot actions with language semantics. However, these methods require complete shapes as input, greatly degrading their effectiveness in real-world scenarios with partial observation.
To address this issue, several approaches that first complete the input shape and then generate grasps based on the completed one are proposed~\cite{kim2025dreamgrasp,iwase2025zerograsp}. Meanwhile, approaches that rely on massive amounts of supervised data can indeed partly overcome the limitations of single-view inputs~\cite{zhong2025dexgraspvla, feng2024dexgangrasp}, but they incur very high data-collection and annotation costs. Moreover, their ability to generalize to unseen object categories or to succeed under extreme occlusions and atypical geometries has yet to be rigorously demonstrated.
Our method improves upon prior work by replacing generic point cloud completion with task-oriented shape completion, substantially enhancing the performance of task-oriented grasp generation from partial point clouds.

\subsection{Masked Autoencoders for Point Clouds}
3D masked autoencoders (MAEs) have recently emerged as a powerful self-supervised paradigm for pre-training on 3D geometric data. By restoring 3D structures from partially masked inputs, these models learn rich representations without explicit supervision. The seminal Point-MAE~\cite{pang2023masked} pioneered this approach for point clouds, establishing the mask and reconstruction pre-training scheme. Subsequent works enhance MAEs for point clouds through various strategies~\cite{zhang2022point,zha2024towards,zhang2024pcp}.
While building upon a similar architecture for shape restoration, our method extends the existing works by introducing an extra discriminative component. This addition enables the identification of implausible input shapes, allowing us to filter incorrect estimations.

\begin{figure*}[!ht]\centering
    \begin{overpic}[width=0.95\linewidth,tics=10]{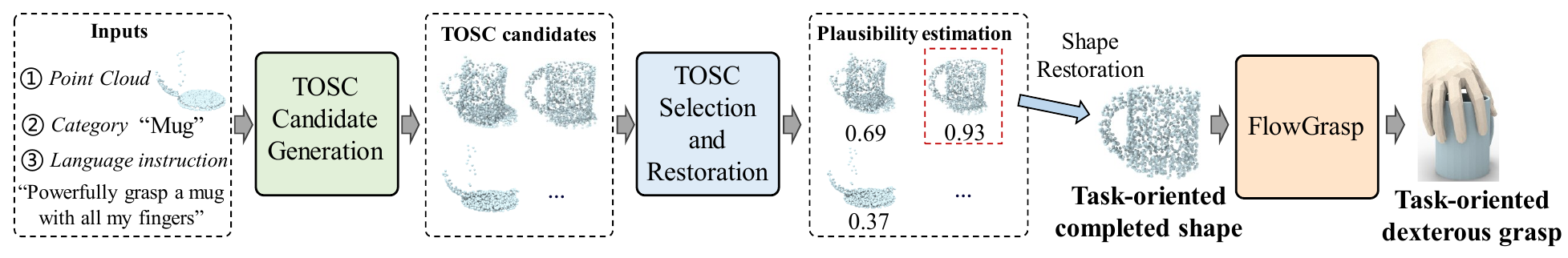}
    \end{overpic}
    \caption{The overview of our method. Taking a partial point cloud of an object, the object's category, and a language description of a manipulation task as input, our method first generates multiple candidates of task-oriented completed shapes by the TOSC candidate generation. It then evaluates the plausibility of the generated candidates and restores the most plausible shape from a global perspective by the TOSC selection and restoration. Last, the task-oriented dexterous grasp is generated by the FlowGrasp.
}
    \label{fig:overall}
\end{figure*}

\begin{figure*}[h]\centering
    \begin{overpic}[width=0.95\linewidth,tics=10]{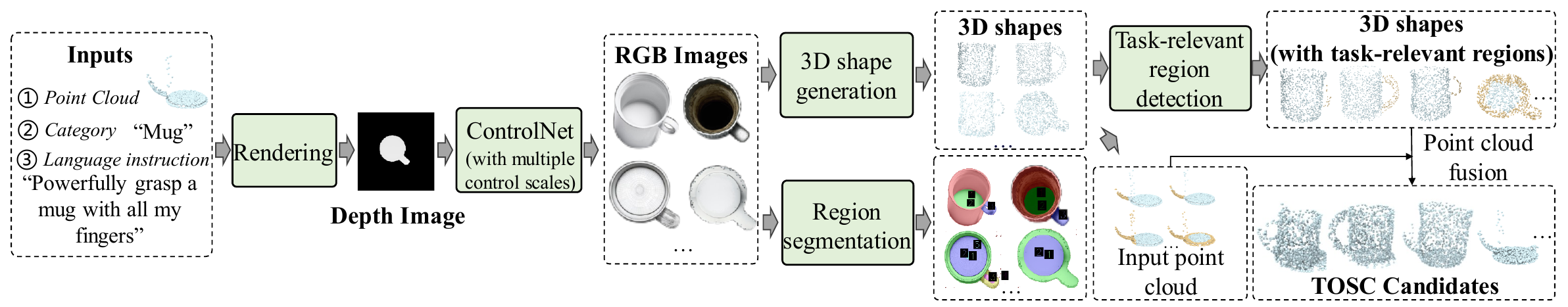}
    \end{overpic}
    \caption{The pipeline of the TOSC candidate generation. First, the input point cloud is rendered into a depth map. Then, the ControlNet is adopted to synthesize multiple RGB images using different control scales. Third, the corresponding 3D shapes are then generated with a 3D shape generation network. After segmenting and detecting task-relevant regions in the generated 3D shapes and input point cloud, a point cloud fusion is performed to generate the TOSC candidates.}
    \label{fig:TOSC_gen}
\end{figure*}
\section{Method}
\label{sec:Method}

\subsection{Overview}
Taking the partial point cloud $P_\text{in}\in \mathbb{R}^{N_\text{in}\times3}$ of a 3D object, its category label $C$, and the language description of a manipulation task $G$ as input, our method completes the shape and generates dexterous grasps as follows. First, it generates multiple candidates of task-oriented completed shapes by leveraging several pre-trained foundation models. Second, it evaluates the plausibility of the generated candidates and optimizes the most plausible shape. Third, the task-oriented grasp is generated by a conditional Flow-Matching grasp generation network. An overview is visualized in Figure~\ref{fig:overall}.

\begin{figure*}[t]\centering
    \begin{overpic}[width=0.95\linewidth,tics=10]{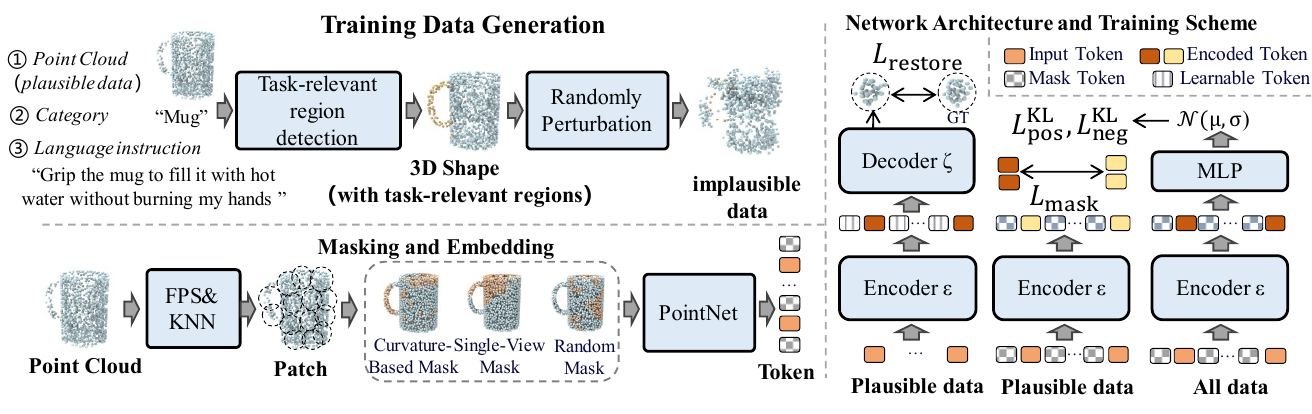}
    \end{overpic}
    \caption{Illustrations of the key components in the TOSC selection and restoration.}
    \label{fig:3DDAE}
\end{figure*}

\subsection{TOSC Candidate Generation}
\label{sec:toscgen}

Despite the recent progress in shape completion, most existing methods complete the geometry based solely on input point clouds, neglecting their relevance to downstream tasks. These methods would yield incorrect estimations, due to the ambiguity caused by data missing. Consequently, grasps generated on these erroneous completions might fail to satisfy the requirements of the specific manipulation tasks.

To alleviate this problem, we opt to task-oriented shape completion, which explicitly conditions shape completion on downstream manipulation tasks. As such, the completion is task-aware, i.e., it focuses on completion of regions that could facilitate the execution of the specific manipulation task, tolerating the imperfections in irrelevant regions.

Direct learning to estimate potential contact regions purely from input point clouds is non-trivial due to the lack of open-world knowledge regarding manipulation tasks.
To solve this problem, we introduce a method that first generates multiple plausible RGB images from the input point cloud by leveraging a pre-trained vision language model. The generated RGB images correspond to multiple plausible shapes, allowing our method to handle ambiguity caused by incomplete observation. The 3D objects corresponding to each RGB image are then generated by using a pre-trained 3D generative model.

Specifically, as shown in Figure~\ref{fig:TOSC_gen}, the input point cloud $P_\text{in}\in \mathbb{R}^{N_\text{in}\times3}$ is rendered into a depth image $I_\text{depth}$. To select the viewpoint for rendering, we adopt the Hidden-Point-Removal~\cite{katz2007direct}, which reformulates viewpoint estimation as a hidden point removal task to generate a viewpoint $V$ that maximizes visible points.
The depth image is then fed into the ControlNet~\cite{zhang2023adding} as a condition to synthesize RGB images, with the object's category $C$ serving as the prompt. 
ControlNet injects $I_\text{depth}$ at multiple U-Net feature levels via zero-initialized convolutional adapters, without requiring any additional task description. 
To generate multiple plausible RGB images, we employ multiple different control scale $\lambda$ on the ControlNet branch to control the balance between strict geometric compliance and semantic completion, where a large $\lambda$ corresponds to ``pay more attention" to the input depth map, preserving fine-grained shape details but also faithfully reproducing any missing-region artifacts, and vice versa.

Having generated the RGB images, 
we then adopt a pre-trained 3D shape generation network~\cite{zhao2025hunyuan3d} $G_M$ to estimate the mesh $M$ for each RGB image, from which a point set $\{P^i_\text{gen}\in \mathbb{R}^{N_\text{gen}\times3}\}_{i=0}^{N_\text{RGB}}$ is sampled.
$G_M$ is a flow-based diffusion model, which adopts double- and single-stream blocks, with DINOv2 Gaint encoding RGB image as condition injection.

The generated shapes are not necessarily perfectly aligned with the input point cloud. Hence, we perform an ICP algorithm to align and fuse the two point clouds. Since our method focuses on the completion of task-relevant regions, we first use SAM~\cite{kirillov2023segment} to segment each RGB image into multiple regions and then adopt a multi-modal large model~\cite{achiam2023gpt} to detect the task-relevant regions in the image space. Those regions are then projected onto both the input point cloud and the generated point cloud to acquire $P^\text{task}_\text{in}$ and $P^\text{task}_\text{gen}$. We optimizing for the scale $k$ and transformation $tr$ via:
\begin{equation}
\underset {k,tr}{\text{argmin}}[\text{CD}(P_\text{in},tr(kP_\text{gen}))+w_\text{task}\text{CD}(P_\text{in}^{\text{task}}, tr(kP^{\text{task}}_{\text{gen}}))],
\label{eq:eq1}
\end{equation}
where $\text{CD}(\cdot,\cdot)$ denotes the chamfer distance, $w_\text{task}$ making the optimization focuses on aligning the task-relevant regions.
We denote the fused point cloud $P_\text{can}\in \mathbb{R}^{N_\text{can}\times 3}$ as a TOSC candidate.

\subsection{TOSC Selection and Restoration}
\label{sec:toscsel}

The generated 3D shape candidates above may be imperfect due to hallucinations during RGB image synthesis and incorrect estimations of shape reconstruction. To solve this problem, we propose a DAE that jointly evaluates the plausibility of each 3D shape candidate and restores the geometry of the most plausible shape from a global perspective. The diagram is shown in Figure~\ref{fig:3DDAE}. 

\subsubsection{Training data generation.}
To effectively train the 3D DAE, we collect a dataset that contains both the plausible and implausible shapes w.r.t the input manipulation task. We define the plausible shapes as those that have sufficient geometry that supports the execution of the input manipulation task. To achieve this, we collect objects from $6$ datasets, i.e. ModelNet40~\cite{wu20153d}, ShapeNetCore~\cite{chang2015shapenet}, ScanObjectNN~\cite{uy2019revisiting}, OmniObject3D~\cite{wu2023omniobject3d}, DexGraspNet~\cite{wang2022dexgraspnet}, and AffordPose~\cite{jian2023affordpose}. In total, the training set of plausible data contains $72,524$ objects, exhibiting significant variations in both synthetic and real-world scenarios.

To generate implausible data, we deliberately sabotage the plausible data by the following steps. We first pair each object with a randomly selected manipulation task and prompt the large language model to identify the contact region implied by the manipulation task. Then, we apply the PartSlip~\cite{liu2023partslip} to identify the task-relevant segment from the object. The implausible data are generated by randomly removing the task-relevant segment, adding noise, and perturbing local patches.

\subsubsection{Network architecture.}
In general, the 3D DAE contains an encoder $\varepsilon(\cdot)$ and an decoder $\zeta(\cdot)$: $l_\text{can}=\varepsilon(P_\text{can}),\hat{P_\text{can}}=\zeta(l_\text{can})$. $P_\text{can}$ is the point cloud of 3D shape candidate, $l_\text{can}$ is the latent vector of $P_\text{can}$, $\hat{P_\text{can}}$ is the restored point cloud.

The encoder $\varepsilon(\cdot)$ consists of $N_\text{encoder}$ standard Transformer blocks. To improve the computational efficiency, we tokenize the input $P_\text{can}$ before feeding it into the encoder $\varepsilon$. Specifically, we first split $P_\text{can}$ into $N_\text{patch}$ local patches by a farthest point sampling (FPS) and a K-nearest neighbor (KNN) algorithm. Each patch is then tokenized by a lightweight PointNet~\cite{qi2017pointnet}.
Inspired by existing works of MAEs~\cite{pang2023masked,wang2021unsupervised,kakogeorgiou2022hide}, during training, we randomly mask several patch tokens in the input $P_\text{can}$ to get mask and visible point cloud $P_\text{can}^\text{mask}$, $P_\text{can}^\text{vis}$. Note that, for the plausible data, we did not mask the patch tokens in the task-relevant segments.

To evaluate the plausibility of the input $P_\text{can}$, we first adopt the latent vector $l_\text{can}$ to estimate the mean $\mu$ and the standard deviation $\sigma$ of a Gaussian distribution $\mathcal{N}(\mu,\sigma)$ by:
\begin{equation}\label{eq:encoder}
\begin{aligned}
\mu&=\text{MLP}_{\mu}(l_\text{can}),\\
\sigma&=\text{MLP}_{\sigma}(l_\text{can}),
\end{aligned}
\end{equation}
where $\text{MLP}_{\mu}(\cdot)$ and $\text{MLP}_{\sigma}(\cdot)$ are fully-connected layers.
Then, a Kullback-Leibler divergence loss $L_\text{pos}^\text{KL}$ is used to optimize the predicted Gaussian distribution $\mathcal{N}(\mu,\sigma)$ of plausible shapes to approximate the standard normal distribution, i.e. $\mathcal{N}(0,1)$, and adopt an $L_\text{neg}^\text{KL}$ to optimize  implausible shapes to approximate another distribution, i.e. $\mathcal{N}(1,1)$.

The decoder $\zeta(\cdot)$ consists of $N_\text{decoder}$ Transformer blocks. It first aggregates features in the tokens and then generates the restored point cloud $P_\text{recon}$ via a fully-connected layer. A chamfer distance loss function between the restored point cloud and the ground-truth is applied:
\begin{equation}\label{eq:rec}
\begin{aligned}
L_\text{restore}= \text{CD}(P_\text{restore},P_\text{GT}),
\end{aligned}
\end{equation}
where $P_\text{GT}$ represents the ground-truth.

Besides, the features of patch tokens are expected to be similar before and after the patch tokens are masked. To improve the robustness under occlusion, we feed all the patch tokens and the masked patch tokens to the encoder $\varepsilon(\cdot)$ to extract features respectively and penalize their differences:
\begin{equation}
L_\text{mask}= \sum_{i=0}^{N_\text{vis}}\text{MSE}[\varepsilon_{T_i}(P_\text{can}^\text{vis}),\varepsilon_{T_i}(P_\text{can})],
\label{eq:p2f}
\end{equation}
where $\text{MSE}[\cdot]$ denotes the mean squared error, $\varepsilon_{T_i}(P_\text{can})$ represents the feature of patch token $T_i$ extracted by encoder $\varepsilon(\cdot)$ taking $P_\text{can}$ as input.
Note that the above $L_\text{recon}$ and $L_\text{mask}$ are only used for positive data. Overall, the training loss function is: $L = L_\text{pos}^\text{KL}+L_\text{neg}^\text{KL}+L_\text{recon}+L_\text{mask}$.

\subsubsection{Network inference.}
During inference, the encoder processes an input point cloud $P_\text{can}$ and outputs a latent vector $l_\text{can}$ as well as a distribution $d_\text{can}=\mathcal{N}(\mu,\sigma)$. The shape plausibility can be estimated with the out-of-distribution likelihood estimation. To achieve this, we consider the KL divergence between the prior and posterior distribution of the input point cloud. The shape plausibility $s_\text{can}$ measured by:
\begin{equation}\label{eq:pla_est}
s_\text{can}=\text{Sigmod}[-\mathcal{D}_\text{KL}(d_\text{can}||\mathcal{N}(0,1)) + \mathcal{D}_\text{KL}(d_\text{can}||\mathcal{N}(1,1))],
\end{equation}
where $\text{Sigmod}[\cdot]$ is the sigmoid function, $\mathcal{D}_\text{KL}$ is the KL divergence. A high $s_\text{can}$ represents that $P_\text{can}$ is closer to the distribution of plausible shapes than that of implausible shapes, indicating $P_\text{can}$ is plausible.
The decoder can then be adopted to generate the restored shape.

\subsection{FlowGrasp}
\label{sec:flowgrasp}
Having a plausible and restored 3D shape, the next step is to compute the dexterous task-oriented grasp. 
Conventional methods typically adopt a conditional variational autoencoder or diffusion models and enforce task and physical constraints by adding weighted penalty terms or by performing gradient-based adjustments at inference time~\cite{wei2024grasp, jiang2021hand}, which either undermines the probabilistic model’s log-likelihood interpretation or incurs significant computational overhead.

\begin{figure}[t]\centering
    \begin{overpic}[width=\linewidth,tics=10]{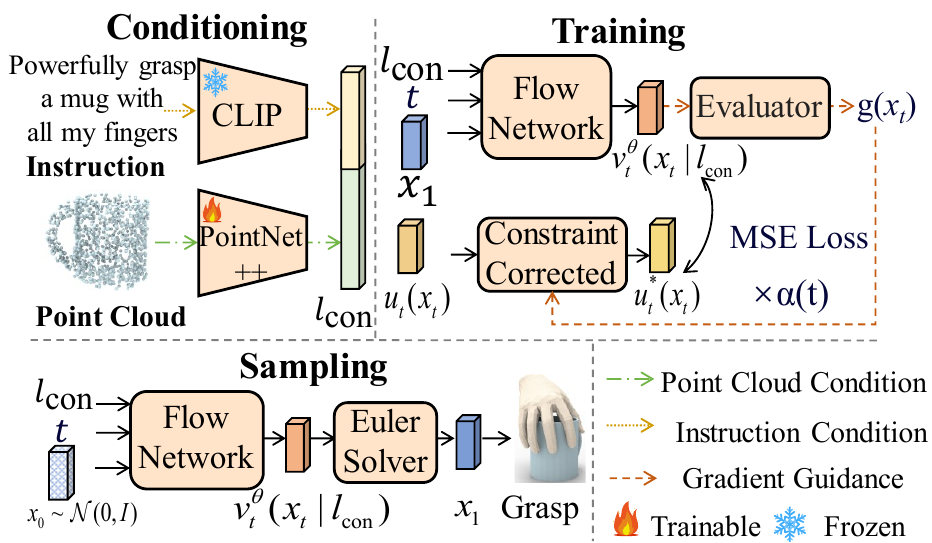}
    \end{overpic}
    \caption{The illustrations of the FlowGrasp.}
    \label{fig:flograsp}
\end{figure}

To solve this problem, we introduce FlowGrasp, a constraint-aware conditional flow matching model that integrates both task and physical constraints directly into training via a single-step, input-side gradient correction—without any extra losses or inference-time adjustments, the illustration is show in Figure~\ref{fig:flograsp}.

Specifically, we form the condition vector $l_\text{con}$ by concatenating the PointNet++ feature of the restored 3D shape with the CLIP embedding of the task language description. During training, we sample a ground-truth grasp vector $x_1$ and an initial noise $x_0\sim \mathcal{N}(0,\text{I})$, pick timestamp $t\in[0,1]$, and interpolate $x_t$ under a fixed diffusion kernel $x_t=(1-t)x_0+t\cdot x_1$.

In the standard flow-matching framework, the instantaneous velocity is $u_t(x_t)=x_1-x_0$. 
To enforce constraints, we apply a one-step correction to this velocity:

 \begin{equation}
u^*_{t}(x_t)=u_t(x_t)-\alpha(t) \nabla(\sum_{i}w_\text{con}^i g_i(x_t)),
\label{eq:speed}
\end{equation}
where each $g_i$ encodes a u or semantic constraint,  $w_\text{con}$ is weighting factor. $\nabla$ is the gradient operator.
$\alpha(t)$ is a time-decay factor.
We then train the network $\theta$ to regress $u^*_{t}(x_t)$ by minimizing:
\begin{equation}
\mathcal{L}_\text{CFM}(\theta)=\mathbb{E}_{x_0,x_1,t,l_\text{con}}||v_t^{\theta}(x_t|l_\text{con})-u_t^{*}(x_t)||^2
\label{eq:eq5}
\end{equation}

During inference, we draw an initial noise vector $x_0\sim \mathcal{N}(0,\text{I})$ and integrate the learned Ordinary Differential Equation (ODE) $dx/dt = v_t^\theta(x_t|c)$ from $t=0$ to $t=1$ to recover the task-oriented dexterous grasp.

\begin{figure*}[t]\centering
    \begin{overpic}[width=\linewidth,tics=10]{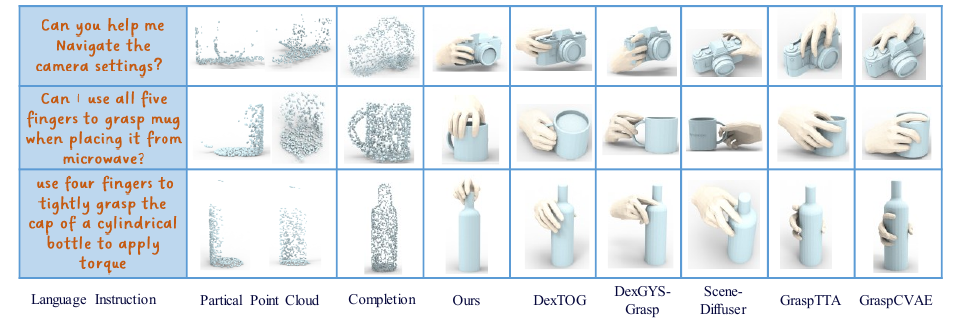}
    \end{overpic}
    \caption{Visual comparison of the generated task-oriented grasps by ours and the baselines.}
    \label{fig:vis_baseline}
\end{figure*}

\begin{table*}[t]
    \centering
    
    \resizebox{0.95\linewidth}{!}{
    \begin{tabular}{c c c c c c c c c c c c}
    \toprule
    \multicolumn{1}{c}{\multirow{2}*{Method}} & \multicolumn{2}{c}{\multirow{1}*{Penetration}} & \multicolumn{2}{c}{\multirow{1}*{Grasp Displace}} & \multicolumn{1}{c}{\multirow{2}*{Contact Radio $\uparrow$}} & \multicolumn{1}{c}{\multirow{2}*{P-FID $\downarrow$}} & \multicolumn{1}{c}{\multirow{2}*{LLM score $\uparrow$}}  &  \multicolumn{3}{c}{\multirow{1}*{Perceptual Score}} \\
    {} & Volume$(cm^3) \downarrow$ & Depth$(cm) \downarrow$ &{Mean$(cm) \downarrow$} & {Var}$(cm) \downarrow$ & {}&{}&{}& SC$\uparrow$& PP$\uparrow $ & IS$\uparrow $  \\ 
    \midrule 
    GraspCVAE~\cite{jiang2021hand}        & 16.84    
    & 0.141 & 3.92 & 4.34 & 94.74$\%$  
    & 39.03 & 55.0 & 1.45 & 1.32 & 1.27 
    \\
    GraspTTA~\cite{jiang2021hand}          & 16.39    
    & 0.159 & 3.71 & 4.19& 96.25$\%$
    & 38.85 & 65.0  & 2.36 & 1.86 & 1.99
    \\
    SceneDiffuser~\cite{huang2023diffusion}& \textbf{6.52}
    & \textbf{0.090} & 3.81 & 4.02 & 
    95.62$\%$
    & 29.38 & 61.7  & 2.27 & 1.95 & 1.82 
    \\
    DexTOG~\cite{zhang2024dextog} & 14.32
    & 0.110 & 3.74 & 3.78 & 96.12$\%$
    & 25.93 & 60.0  & 2.54 & 2.22 & 2.13 
    \\
    DexGYSGrasp~\cite{wei2024grasp} & 7.16
    & 0.096 & 3.76 & 3.99 & 97.20$\%$
    & 25.98 & 68.3  & 3.04 & 2.23 & 1.99 
    \\
    \midrule
    Ours    & 6.87 & \textbf{0.090}   & $\textbf{3.11}$ & $\textbf{3.54}$ & $\textbf{98.30}\%$
    & $\textbf{21.60}$ & $\textbf{88.3}$   & $\textbf{4.38}$ & $\textbf{3.84}$ & $\textbf{3.80}$ 
    \\
    \toprule
    \end{tabular}}
    \caption{Quantitative comparisons of task-oriented grasping on the OakInk-PartialPC dataset.}
    \label{tab:performance}
\end{table*}

\begin{table}[t]
    
    \centering
    \resizebox{\linewidth}{!}{
    \begin{tabular}{c c c c}
    \toprule
    Method & CD-$\ell{_2} \times 10^{-4}\downarrow$& F-Score@1$ \uparrow$& DCD$\downarrow$
    \\ 
    \midrule
   PointAttn~\cite{wang2024pointattn} 
     & 4.58 & 0.512 & 0.698 \\
    SVDFormer~\cite{zhu2023svdformer}
    & 3.71 & 0.643 & 0.603\\
    SymmCompletion~\cite{yan2025symmcompletion}
    & 3.94 & 0.618 & 0.611 \\
    \midrule
    Ours
     & \textbf{1.66} & \textbf{0.860} & \textbf{0.488}\\
    \toprule
    \end{tabular}}
    \caption{Quantitative comparisons of point cloud completion on the OakInk-PartialPC dataset.}
    \label{tab:com}
\end{table}

\section{Experiments}
\label{sec:Experiments}

\subsection{Implementation Details}
The number of points in both the partial and candidate clouds $N_\text{in}$ and $N_\text{can}$ is 2048. We convert the depth image to RGB images with ControlNet~\cite{zhang2023adding} and then reconstruct point clouds from RGB using Hunyuan3D-DiT-v2-mini-Fast~\cite{zhao2025hunyuan3d}. Task-relevant region detection is achieved by GPT-4o. FlowGrasp is trained on a single NVIDIA GeForce RTX 4090 GPU for 350 epochs with the Adam optimizer and a batch size of 64. The 3D DAE is trained for 300 epochs with a learning rate of 0.0005 and a weight decay of 0.05. 

\subsection{Experimental Dataset}
Our method, except for the 3D DAE, is trained on a new dataset named OakInk-PartialIPC, which containing partial point clouds created from OakInk~\cite{yang2022oakink}. To generate the partial point clouds, we uniformly sample viewpoints around each object and render depth images with random scanning noise and random foreground occlusion to simulate depth maps scanned in real-world environments. The corresponding language instructions are drawn from the CapGrasp dataset~\cite{li2024semgrasp}, which provides task descriptions specifically tailored to OakInk.

\subsection{Evaluation Metrics}
To comprehensively evaluate the quality of the generated grasps, we employ multiple evaluation metrics: 1) \emph{Penetration} quantifies the volumetric overlap and maximum insertion depth between the grasp and object~\cite{hasson2019learning}; 2) \emph{Grasp displacement} evaluates the grasping stability by measuring the displacement of object's mass center under external forces for each grasp; 3) \emph{Contact ratio} reflects the quality of hand–object interaction by calculating sample-level contact ratio~\cite{tzionas2016capturing}; 4) \emph{Point cloud Fréchet Inception Distance (P-FID)} compares high-level feature distributions of generated grasps against ground truth~\cite{nichol2022point}; 5) \emph{LLM score} evaluates the quality of grasps with GPT4v~\cite{li2024semgrasp}; 6) \emph{Perceptual Scores} are acquired by a user study in which human evaluators each assess $100$ distinct grasps and assign scores from 0 (poor) to 5 (excellent) on three criteria: Semantic Consistency (SC), Physical Plausibility (PP), and Interaction Stability (IS).

For evaluation of point cloud completion, we use the standard metrics: 1) Chamfer Distance (CD)~\cite{wang2024pointattn}; 2) F1-score~\cite{yan2025symmcompletion} ; 3) Density-aware CD (DCD)~\cite{wu2021balanced}.

\begin{figure*}[t]
  \centering
   \includegraphics[width=0.95\linewidth]{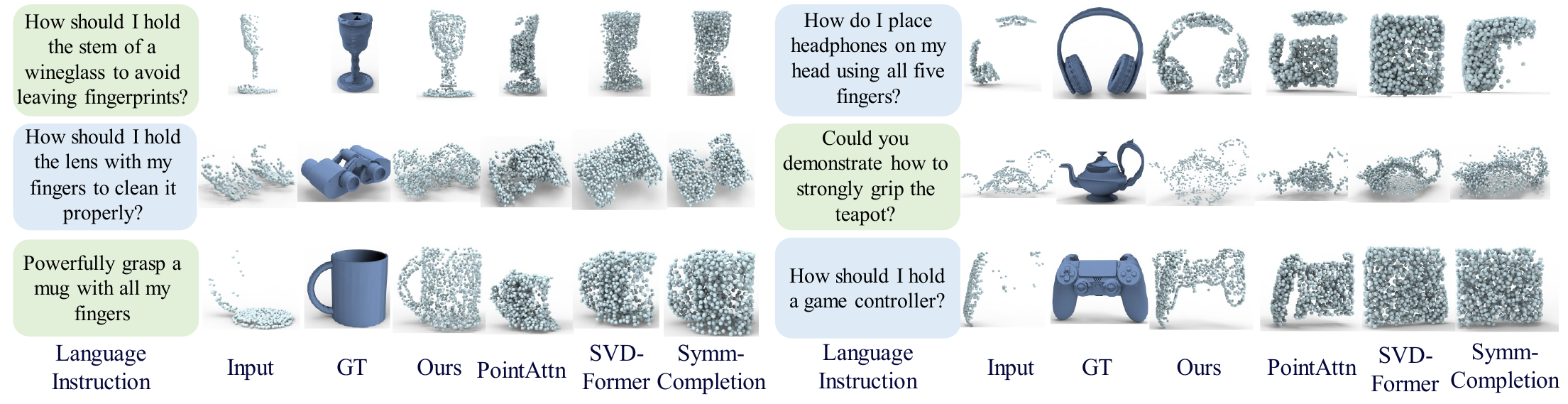}
   \caption{Visual comparisons of point cloud completion. }
   \label{fig:complete}
\end{figure*}

\begin{table*}[t]
    \centering
    \resizebox{0.95\linewidth}{!}{
    \begin{tabular}{c c c c c c c c c c c c}
    \toprule
    \multicolumn{1}{c}{\multirow{2}*{Method}} & \multicolumn{2}{c}{\multirow{1}*{Penetration}} & \multicolumn{2}{c}{\multirow{1}*{Grasp Displacement}} & \multicolumn{1}{c}{\multirow{2}*{Contact Radio $\uparrow$}} & \multicolumn{1}{c}{\multirow{2}*{P-FID $\downarrow$}} & \multicolumn{1}{c}{\multirow{2}*{LLM score $\uparrow$}}  & \multicolumn{3}{c}{\multirow{1}*{Perceptual Score}} \\
    {} & Volume$(cm^3) \downarrow$ & Depth$(cm) \downarrow$ &{Mean$(cm) \downarrow$} & {Var}$(cm) \downarrow$ & {}&{}&{}& SC$\uparrow$& PP$\uparrow $ & IS$\uparrow $ \\ 
    \midrule
    \emph{ w/o TCG}
    & 6.70 & 0.098 & 3.50 & 3.79  & 97.74$\%$
    & 22.34 & 71.7  & 2.63 & 0.81 & 1.72 \\
    \emph{w/o TSR}
    & 7.96 & 0.093 & 3.35 & 3.68  & 98.84$\%$
    & 22.96 & 75.0  & 2.36 & 2.63 & 2.72 \\
    \emph{w/o TOSC}
    & 7.12 & 0.090 & 3.51 & 3.94  & 96.79$\%$
    & 23.83 & 66.7   & 2.18 & 1.36 & 2.18 \\
    \emph{w/o token masking} 
    & 7.58 & 0.087 & 3.21 & 3.71 & \textbf{99.50}$\%$
    & 22.53 & 78.3   & 3.09 & 3.18 & 3.36 \\
    \emph{w/o gradient guidance}
    & 6.78 & \textbf{0.086} & 3.43 & 3.70  & 98.83$\%$
    & 24.01 & 83.3  & 3.72 & 2.18 & 3.63 \\
    \midrule
    The full method
    & $\textbf{6.87}$ & 0.090               & $\textbf{3.11}$ & $\textbf{3.54}$ & $98.30\%$
    & $\textbf{21.60}$ & $\textbf{88.3}$   & $\textbf{4.38}$ & $\textbf{3.84}$ & $\textbf{3.80}$\\
    \toprule
    \end{tabular}}
    \caption{Ablation studies of several crucial components in our method.}
    \label{tab:abs}
\end{table*}

\begin{table}[t]
    \centering
    \resizebox{\linewidth}{!}{
    \begin{tabular}{c c c c c c}
    \toprule
     \multicolumn{1}{c}{\multirow{2}*{Method}} & \multicolumn{2}{c}{\multirow{1}*{Penetration}} & \multicolumn{2}{c}{\multirow{1}*{Grasp Displace}} & \multicolumn{1}{c}{\multirow{2}*{P-FID $\downarrow$}} \\
    {} & Volume$(cm^3) \downarrow$ & Depth$(cm) \downarrow$ &{Mean$(cm) \downarrow$} & {Var}$(cm) \downarrow$ & {}  \\ 
    \midrule
   GraspCVAE  
     & 16.02 & 1.18 & 4.59 & 4.77 & 57.01 \\
    GraspTTA
    & 14.42 & \textbf{1.04} & 4.59 & 4.83 & 53.58 \\
    SceneDiffuser  
    & 12.43 & 1.16 & 4.53 & 4.70 & 46.51 \\
    DexTOG  
    & 18.72 & 1.31 & 4.42 & 4.80 & 50.53 \\
    DexGYSGrasp
    & 12.60 & 1.28 & 4.77 & 4.54 & 46.42  \\
    \midrule
    Ours
     & \textbf{11.53} & 1.32 & \textbf{4.21} & \textbf{4.52} & \textbf{42.97}\\
    \toprule
    \end{tabular}}
    \caption{Quantitative comparisons of task-oriented grasping on novel category.}
    \label{tab:novel}
\end{table}

\subsection{Compare to Baselines }
We first compare our method to baselines in task-oriented dexterous grasping.
Table \ref{tab:performance} reports the quantitative comparisons against several baselines on the OakInk-PartialPC dataset. It shows that our method achieves state-of-the-art performances in metrics of Grasp Displace and competitive performances in metrics of Penetration, demonstrating the superior physical reliability of the generated grasps. Note that the lower penetration volume of SceneDiffuser~\cite{huang2023diffusion} stems from their conservative strategy, which keeps unnecessary distances between hand and object. Our method, in contrast, maintains better balances between penetration avoidance and grasp stability.
Moreover, our method achieves the highest LLM Score and perceptual scores, showing its ability to produce high-quality semantically correct grasps that fulfill the language instructions. 
Qualitative comparisons are visualized in Figure~\ref{fig:vis_baseline}.

We then evaluate our method in terms of task-oriented shape completion on the OakInk-PartialPC dataset. Table \ref{tab:com} reports quantitative comparisons with several baselines for generic point cloud completion. The baselines are recent state-of-the-art methods, including PointAttn~\cite{wang2024pointattn}, SVDFormer~\cite{zhu2023svdformer}, and SymmCompletion~\cite{yan2025symmcompletion}. Our method outperforms all baselines by a large margin in all metrics, revealing the advantages of task-oriented shape completion. Qualitative comparisons are shown in Figure \ref{fig:complete}.

\subsection{Evaluation of Zero-shot Generality}
We first evaluate the zero-shot generalization in terms of handling novel object categories and novel language instructions. We experiment on $9$ novel object categories. Each category contains about $100$ objects and several novel language instructions. As shown in Table~\ref{tab:novel}, our method achieves the best performance across all metrics. It demonstrates that our method can robustly handle novel categories and untrained language instructions in open-world scenarios, thanks to our design of using pre-trained foundation models.

\subsection{Ablation Studies}

\subsubsection{TOSC Candidate Generation (TCG).}
TCG generates task-oriented shape completion candidates, which provide zero-shot generation capability to our method. To evaluate its impact, we conduct an ablation study by removing TCG and feeding the input partial point cloud to the TOSC restoration as the only candidate. As shown in Table~\ref{tab:abs}, Results indicate that this removal significantly degrades performance, confirming its necessity.

\subsubsection{TOSC Selection and Restoration (TSR).}
TSR selects the plausible shape and restores the shape. To evaluate its effect, we directly select the candidate with the minimum chamfer distance to the input point cloud and remove the shape restoration process. The inferior performance demonstrates its significance.

\subsubsection{Task-Oriented Shape Completion (TOSC).}
We replace the task-oriented shape completion with a generic shape completion~\cite{wang2024pointattn}. The resulting performance drop confirms our idea of task-oriented shape completion.

\subsubsection{Token Masking.}
The token masking scheme of the 3D DAE is proposed to improve the robustness. We remove this component and evaluate the performance. This variant exhibits a performance drop, confirming its effectiveness.

\subsubsection{Gradient Guidance.}
Gradient guidance constrains the model to generate grasping actions that satisfy semantic and geometric consistency while optimizing the velocity field. To evaluate its necessity, we replace it with a direct loss-based optimization. This variant underperformed the full method, confirming the superiority of gradient guidance.

\section{Conclusion}
\label{sec:Conclusion}
We have studied task-oriented shape completion, a new task that focuses on completing the potential contact regions rather than the entire shape. To achieve this, we propose a method that first generates multiple task-oriented shape completion candidates and then selects the most plausible one as well as optimizes its geometry by learning a 3D discriminative autoencoder. This method achieved state-of-the-art performance on both task-oriented dexterous grasping and task-oriented shape completion. In particular, it produces high-quality results in challenging scenarios.

\section{Acknowledgments}
This work was supported by the Natural Science Foundation of Hunan Province (2023JJ20051), the Science and Technology Innovation Program of Hunan Province (2023RC3011), the Cornerstone Foundation of NUDT (JS24-03) and the National Natural Science Foundation of China (62472434).
\bibliography{aaai2026}

\end{document}